\documentclass[letterpaper, 10 pt, conference]{ieeeconf}  

\IEEEoverridecommandlockouts                              

\usepackage{graphics} 
\usepackage{epsfig} 
\usepackage{mathptmx} 
\usepackage{times} 
\usepackage{amsmath} 
\usepackage{amssymb}  

\usepackage[]{algorithm2e}

\usepackage{multirow}
\usepackage{makecell}

\renewcommand{\eqref}[1]{\mbox{Eq. ~\ref{#1}}}
\usepackage{graphicx}
\usepackage{multirow}
\newcommand{\mypar}[1]{\vspace{0.3cm}\noindent\textbf{#1}}

\newcommand{\ie}{\textit{i}.\textit{e}.~}

\usepackage{mathtools}
\usepackage{colortbl}
\definecolor{lightsteelblue}{RGB}{176,196,222}
\definecolor{lightsteelred}{RGB}{230,176,160}
\definecolor{lightsteellila}{RGB}{175,181,224}
\definecolor{lightsteelgreen}{RGB}{182,214,207}
\definecolor{tablecol2}{RGB}{204, 204, 255}
\definecolor{tablecol1}{RGB}{255, 217, 179}

\usepackage[usenames,dvipsnames]{xcolor}

\makeatletter
\def\footnoterule{\kern-3\p@
  \hrule \@width 2in \kern 2.6\p@} 
\def\endthebibliography{%
	\def\@noitemerr{\@latex@warning{Empty `thebibliography' environment}}%
	\endlist
}
\makeatother

\title{\LARGE \bf
A Comparative Analysis of Decision-Level
Fusion for \\Multimodal Driver Behaviour Understanding}

\author{ Alina Roitberg  \quad\quad Kunyu Peng \quad\quad Zdravko Marinov \\ 
Constantin Seibold \quad\quad\quad David Schneider \quad\quad\quad Rainer Stiefelhagen 
	\\
\\Institute for Anthropomatics and Robotics
\\ Karlsruhe Institute of Technology
\\  {\tt\small \{firstname.lastname\}@kit.edu}
}

\usepackage{todonotes}		
\usepackage{booktabs} 		
\usepackage{amsfonts}		
\usepackage{tikz}

\newcommand\copyrighttext{%
	\footnotesize Accepted at IV 2022, © IEEE. Personal use is permitted, but republication/redistribution requires IEEE permission.  Permission from IEEE must be obtained for all other uses, in any current or future media,including reprinting/republishing this material for advertising or promotional purposes, creating new collective works, for resale or redistribution to servers or lists, or reuse of any copyrighted component of this work in other works.}

\newcommand\copyrightnotice{%
	\begin{tikzpicture}[remember picture,overlay]
	\node[anchor=south,yshift=10pt] at (current page.south) {\fbox{\parbox{\dimexpr\textwidth-\fboxsep-\fboxrule\relax}{\copyrighttext}}};
	\end{tikzpicture}%
}
\begin{document}

\maketitle
\copyrightnotice{}
\thispagestyle{empty}
\pagestyle{empty}


\begin{abstract}

Visual recognition inside the vehicle cabin leads to safer driving and more intuitive human-vehicle interaction but such systems face substantial obstacles as they need to capture different granularities of driver behaviour while dealing with highly limited body visibility and changing illumination. 
\textit{Multimodal} recognition mitigates a number of such issues: prediction outcomes of different sensors complement each other due to  different modality-specific strengths and weaknesses.
While several late fusion methods have been considered in previously published frameworks, they constantly feature different architecture backbones and building blocks making it very hard to isolate the  role of the chosen late fusion strategy itself.

This paper presents an empirical evaluation of different  paradigms for decision-level late  fusion in video-based driver observation.
We compare seven different mechanisms for joining the results of single-modal classifiers which have been both popular, (e.g. score averaging) and not yet considered (e.g. rank-level fusion) in the context of driver observation evaluating them based on different criteria and  benchmark  settings.
This is the first systematic study of strategies for fusing outcomes of multimodal predictors inside the vehicles, conducted with the goal to provide guidance for fusion scheme selection.

\end{abstract}


\section{Introduction and Related Work}

\begin{figure}[!t]
	\centering
	\includegraphics[width=\linewidth]{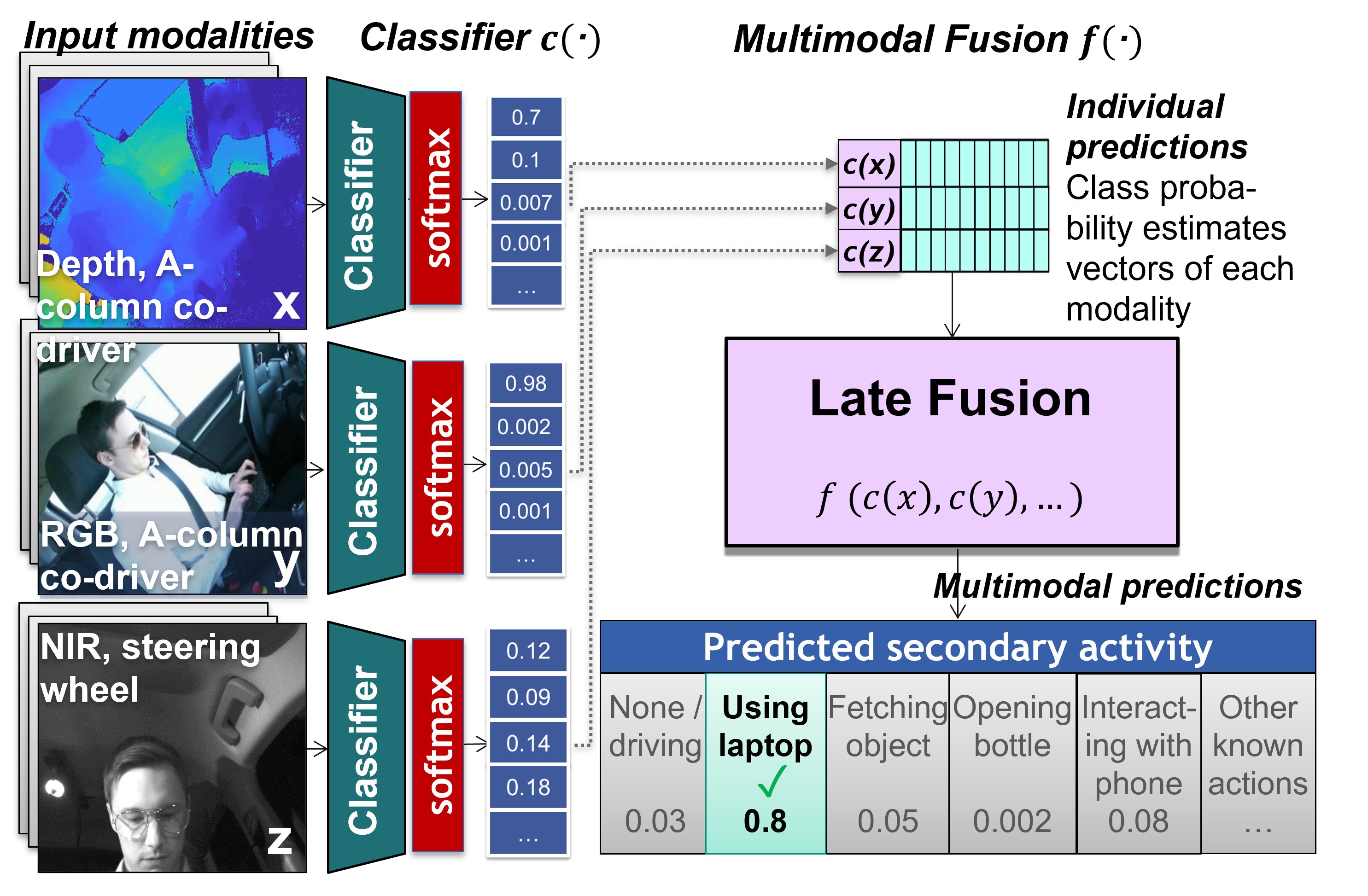}
	\caption{
		A high-level overview of a multimodal driver observation framework featuring three separate classification streams, with their fusion is carried out after the  single-modal predictions were obtained.
		We implement and  study different techniques for linking such single-modal outcomes.
}
	\label{fig:overview}
\vspace{-0.4cm}
\end{figure}

\begin{figure*}[!t]
	\centering
	\includegraphics[width=\linewidth]{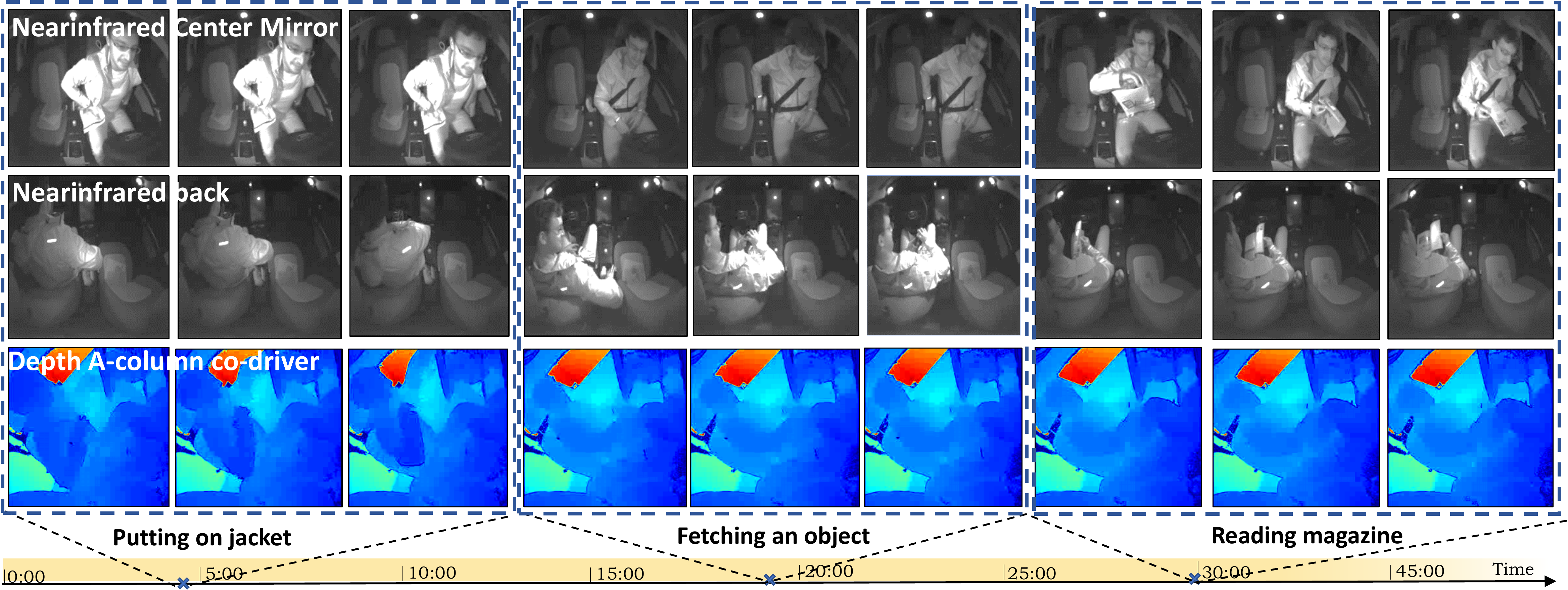}
	\caption{
		Example of a multimodal driver activity recognition setting  with highly distractive behaviours during automated driving. Different modalities have their specific strengths and limitations depending on the  visible portion of the  cabin and sensor-specific characteristics. For example, both RGB and NIR cameras might capture unnecessary textures  constituting additional noise, while RGB sensors depend on the illumination. Depth data is less sensitive to illumination changes and skips unnecessary texture details (e.g., clothing) but might also miss details important for the behaviours-of-interest.
}
	\label{fig:modality_examples}
\vspace{-0.4cm}
\end{figure*}

\textit{Multimodality} increasingly gains attention in driver observation systems~\cite{kopuklu2021driver,MartinRoitberg2019,Jain2015a, khan2021modified}: prediction outcomes of multiple sensors complement each other due to  modality-specific strengths and weaknesses as well as different visibility (examples in Figures \ref{fig:overview} and \ref{fig:modality_examples}).
Rising levels of automation increase human freedom, leading to drivers being engaged in distractive behaviours more often while the type of activities become increasingly diverse.
This is very challenging for \textit{unimodal} driver observation systems, which need to capture different complexities and granularities of situations inside the cabin despite strongly restricted body visibility.
For example, frameworks developed for manual driving often focus on the face view to capture the  attentiveness regarding the driving scene~\cite{Jain2015a, rangesh2020driver, ohn2014hand}.
However, as the driver is gradually relieved from actively steering the car, activities such as \textsl{working on laptop} or \textsl{reading magazine}, which were almost unthinkable until now, become more common.
Equipping the vehicle with multiple complementing sensors enables recognition of very different behaviour types, but \textit{how to link the information} becomes an important question.

The state-of-the-art of multimodal driver activity recognition constantly changes depending on different architecture choices, losses and classifier components~\cite{kose2019real,chen2020driver,martin2018body,MartinRoitberg2019,kopuklu2021driver,wharton2021coarse}, but a large portion of such methods employ late fusion via score averaging to link the information~\cite{MartinRoitberg2019, kopuklu2021driver, martin2018body, khan2021modified}.
Multimodal fusion algorithms can be grouped depending on the point of fusion, (\textit{e.g.}, early-, mid-, or late-fusion) and based on the methodology (learning- and  decision-based approaches). 
The learning-based approaches \textit{learn} to combine the streams (and can therefore be applied at different information processing stages). 
In \textit{decision-level fusion}, on the other hand, individual unimodal probability estimates for each behaviour category are obtained a priori, after which a transformation function, such as average, product, or voting joins them into a common multimodal decision.

This work conducts the first systematic study of strategies for fusing outcomes of multimodal predictors at decision-level for visual recognition inside the vehicle cabin.
Despite omitting intra-modality correlations at earlier stages, decision-level operations bring important advantages. 
First, in contrast to the learning-based methods, the multimodal systems operating on decision-level  are highly modular, as the individual modalities with pretrained classifiers can be flexibly plugged-in or removed without any additional retraining.
As a consequence, if one of the sensors is damaged, a decision-level fusion system would simply exclude it from contributing, whereas training of the fusion model would need to be revisited in standard  learning-based approaches, as the feature vector appearance changes.

Among  decision-level fusion techniques, averaging of the obtained \textit{Softmax} 
scores  is presumably the most common choice in driver activity recognition~\cite{MartinRoitberg2019, kopuklu2021driver, martin2018body, khan2021modified}.
In the broader fields of general machine learning and computer vision, this approach is also highly popular~\cite{ye2015temporal,imran2016human,baradel2017human,ardianto2018multi,dawar2018convolutional,carreira2017quo,dhiman2020view,cai2021jolo}, but other strategies, which were rather overlooked in the field of driver observation, such as the max rule \cite{kamel2018deep, ardianto2018multi, dhiman2020view, rani2021kinematic} or the product rule \cite{imran2016human, wang2017scene, kamel2018deep, Pushpajit2018, dawar2018data, zhao20193d, wei2019fusion, dhiman2020view, rani2021kinematic} also gained  attention. A theoretical study of such methods from the pre-Deep Learning era is provided in \cite{kittler1998combining}.
Rank-level decision-level fusion, such as Borda Count voting \cite{emerson2013original,Erp2002, ramanathan2019combining} and Reciprocal Rank Voting \cite{cormack2009reciprocal} are less popular, but have been successfully applied in the field of biometric identification \cite{damer2017general, sharma2015rank}.
There are several works targeting multimodal fusion through learning-based methods,\textit{ e.g.}, using SVM, LSTM, or neural network fusion layers~\cite{ohn2014head,jain2016brain4cars,wang2016exploring,kose2019real,roitberg2019analysis,kazakos2019epic}. We, however, consider these out of the scope of this work, as these require additional training and cannot be directly used out-of-the-box for target fusion at the decision-level. 
Nevertheless, recent computer vision research is rather  focused on the generation of high performing single-modal classifiers while fusion strategies are considered of lesser importance and few of them are systematically explored in combination with novel CNN-based methods.

\mypar{Summary and contributions}
In this work, our goal is to implement and systematically evaluate different strategies for decision-level fusion in the context of multimodal driver behaviour assessment.
We build upon recent advances in driver observation and train a neural network often utilized in this task separately for each of the eight modalities of a standard multimodal driver activity recognition testbed~\cite{MartinRoitberg2019}. 
We compare 10 different mechanisms for joining the results of single-modal classifiers which have been both popular, (\textit{e.g.}, score averaging) and not yet considered, (\textit{e.g.}, rank-level fusion) in the context of driver observation and evaluate them based on different criteria and  benchmark  settings.
Our results indicate that the choice of fusion mechanisms  impacts the model performance. Furthermore, the commonly employed average-fusion being outperformed by several other methods in all evaluation settings and metrics.
Of the considered methods, product-fusion and max-fusion yielded the best recognition results. Interestingly, while max-fusion oftentimes outperformed product-fusion by a small margin, product-fusion is consistently more effective when it comes to top-5 accuracy, indicating, that it might be useful in coarser recognition. We further compare our multimodal system to the best performing unimodal view.
Overall, multimodality is clearly beneficial for almost all behaviour types, but the effect depends on visibility and recognition difficulty: the largest benefits of multimodality were observed in driver behaviours with medium recognition difficulty.
To the best of our knowledge, this is the first systematic study of strategies for decision-level fusion inside the vehicle cabin.
Our experiments provide empirical evidence that the commonly employed late fusion via  averaging is not the most effective way of linking unimodal driver observation results, and we hope that our study will provide guidance for better fusion scheme selection in the future.

\section{Revisiting Late Fusion for Video-based Driver Observation} 

In this paper, we analyze different approaches for fusing the decision-level predictions of multiple visual driver observation models.
That is, given $N$ different modalities with inputs $x_i, i \in {1...N}$ (see examples in Figure ~\ref{fig:modality_examples}) and $N$ pretrained unimodal classifiers with predictions $c_i(x_i), i \in {1...N}$ containing probability estimates for each category, our goal is to correctly identify the potentially distractive behaviour of the driver by linking the information of these different modalities effectively. 
To this intent, we employ the I3D  architecture~\cite{carreira2017quo} as the unimodal classifiers backbone, which has shown excellent results in driver activity recognition~\cite{MartinRoitberg2019, roitberg2020cnn}. 
We train the models for each modality individually. Afterwards, we utilize different variants of the decision-level fusion module which takes multiple class probability estimates produced by the individual classifiers as input and joins them to reach the final multi-modal decision. Note, that we specifically target \emph{decision-level} approaches that do not require any architecture training or changes in architecture. 
While multiple introduced approaches address multimodal fusion with learning-based methods~\cite{ohn2014head,jain2016brain4cars,wang2016exploring,kose2019real}, such approaches are out of the scope of this work.
In total, we implement 
seven different strategies for multimodal decision-level fusion, which we now discuss in detail. 


\subsection{Score-level fusion}
In score-level fusion, the goal is to combine the predictions of $N$ classifiers on a $d$-classification task based on their class probability estimates $c(x_i)$, where $i \in \{1...N\}$. We investigate fusing the predictions $c(x_i)$ via summation or averaging, maximum, and product of the probability vectors. For this, we  introduce the following notation in Table   \ref{tbl:notation}:

\begin{table}[!h]
	\begin{center}
		\resizebox{0.5\textwidth}{!}{%
			\begin{tabular}{@{}|l| l|}
				\hline
				$N \in \mathbb{N}$ & Number of classifiers. \\ 
				$d \in \mathbb{N}$ & Number of classes. \\ 
				$c(x_i) \in \mathbb{R}^d$ & Probability estimates of $i^{\text{th}}$ classifier. \\ 
				$c(x_i)_j \in \mathbb{R}$ & Probability estimate of $i^{\text{th}}$ classifier for $j^{\text{th}}$ class. \\ 
				$c(X):=\{c(x_1)...c(x_N)\} \in \mathbb{R}^{d \times N}$ & Set of all probability estimates. \\ 
				$c(X)_j:=[c(x_1)_j...c(x_N)_j] \in \mathbb{R}^N$ & Predictions for $j^{\text{th}}$ class from all classifiers. \\ 
				$r_{ij}$ & Rank of $j^{\text{th}}$ class in $c(x_i)$. \\ \hline
			\end{tabular}  
		}
	\end{center}
	\caption{Notation for all the late fusion equations.}
	\label{tbl:notation}
\end{table}
Note that the fusion results from all the methods we investigate can be used in combination with $argmax(\cdot)$ to produce the final class prediction.

\mypar{Sum-fusion and score averaging:}
The sum-fusion (often referred to as average-fusion) $f_{SUM}(\cdot)$ for $N$ classifiers is defined as:
\begin{equation}
    f_{SUM}(c(X))=\frac{1}{N} \sum_{i=1}^{N} c(x_i)
\end{equation}
Note that the division by $N$ does not change the ranking of the summed predictions, but serves to regularize the output to sum up to 1. 
This fusion strategies has presumably been the most popular choice for fusion at decision-level in driver observation~\cite{MartinRoitberg2019, kopuklu2021driver, martin2018body, khan2021modified}.

\mypar{Median-based fusion:} The median-fusion $f_{MED}(\cdot)$ for $N$ classifiers is defined as:
\begin{equation}
    f_{MED}(c(X))= [med(c(X)_1)...med(c(X)_{d})]
\end{equation}
where
\begin{equation}
    med(x)=
    \begin{cases}
    \hat{x}_{(d+1)/2},& \text{if } d \text{ is odd}\\
    \frac{1}{2}(\hat{x}_{(d/2)} + \hat{x}_{(d/2) + 1}),              & \text{otherwise} 
    \end{cases}
\end{equation}
Here $\hat{x}_d$ is defined as the $d^{\text{th}}$ element of $\hat{x}$ and $\hat{x}$ is $x$ sorted in ascending order.

\mypar{Max-fusion:} The max-fusion $f_{MAX}(\cdot)$ for $N$ classifiers is defined as:
\begin{equation}
    f_{MAX}(c(X))= [max(c(X)_1)...max(c(X)_d)]
\end{equation}

. 
\setlength{\tabcolsep}{4.5pt}

\begin{table*}[ht]
    \centering
\scalebox{0.95}{\begin{tabular}{cccccccccccccccc} 
\toprule
 & \multirow{3}{*}{\begin{tabular}[c]{@{}c@{}}\textbf{Fusion}\\\textbf{Method }\end{tabular}} & \multicolumn{4}{c}{\textbf{\#Mod=2 }} &  & \multicolumn{4}{c}{\textbf{\#Mod=4 }} &  & \multicolumn{4}{c}{\textbf{\#Mod=8 }} \\ 
\cline{3-6}\cline{8-11}\cline{13-16}
 &  & \multicolumn{2}{c}{\textbf{Balanced Acc. }} & \multicolumn{2}{c}{\textbf{Unbalanced Acc. }} &  & \multicolumn{2}{c}{\textbf{Balanced Acc. }} & \multicolumn{2}{c}{\textbf{Unbalanced Acc. }} &  & \multicolumn{2}{c}{\textbf{Balanced Acc. }} & \multicolumn{2}{c}{\textbf{Unbalanced Acc.}} \\
 &  & \textbf{\textcolor{red}{Top-1 }} & \textbf{Top-5 } & \textbf{\textcolor{red}{Top-1 }} & \textbf{Top-5} &  & \textbf{\textcolor{red}{Top-1 }} & \textbf{Top-5 } & \textbf{\textcolor{red}{Top-1 }} & \textbf{Top-5} &  & \textbf{\textcolor{red}{Top-1 }} & \textbf{Top-5 } & \textbf{\textcolor{red}{Top-1 }} & \textbf{Top-5 } \\ 
\midrule
\multirow{6}{*}{\rotatebox[origin=c]{90}{\textbf{Score-level}}} & \textbf{Avg.} w/o weight. (standard)& {\cellcolor[rgb]{0.867,0.867,0.871}}47.81 & 70.49 & {\cellcolor[rgb]{0.867,0.867,0.871}}42.57 & 67.71 &  & {\cellcolor[rgb]{0.867,0.867,0.871}}51.44 & 77.88 & {\cellcolor[rgb]{0.867,0.867,0.871}}46.06 & 75.41 &  & {\cellcolor[rgb]{0.867,0.867,0.871}}54.69 & 80.9 & {\cellcolor[rgb]{0.867,0.867,0.871}}49.72 & 78.72 \\
 & \textbf{Average} w. weight. & {\cellcolor[rgb]{0.867,0.867,0.871}}47.52 & 70.49 & {\cellcolor[rgb]{0.867,0.867,0.871}}42.2 & 67.71 &  & {\cellcolor[rgb]{0.867,0.867,0.871}}51.46 & 77.88 & {\cellcolor[rgb]{0.867,0.867,0.871}}46.24 & 75.41 &  & {\cellcolor[rgb]{0.867,0.867,0.871}}54.96 & 80.75 & {\cellcolor[rgb]{0.867,0.867,0.871}}50.09 & 78.72 \\
 & \textbf{Median} & {\cellcolor[rgb]{0.867,0.867,0.871}}47.81 & 70.49 & {\cellcolor[rgb]{0.867,0.867,0.871}}42.57 & 67.71 &  & {\cellcolor[rgb]{0.867,0.867,0.871}}51.87 & 80.07 & {\cellcolor[rgb]{0.867,0.867,0.871}}46.79 & 78.35 &  & {\cellcolor[rgb]{0.867,0.867,0.871}}54.01 & 84.62 & {\cellcolor[rgb]{0.867,0.867,0.871}}49.54 & 81.83 \\
 & \textbf{Max} & {\cellcolor[rgb]{0.867,0.867,0.871}}47.52 & 70.19 & {\cellcolor[rgb]{0.867,0.867,0.871}}42.2 & 66.97 &  & {\cellcolor[rgb]{0.867,0.867,0.871}}\textbf{53.26} & 77.45 & {\cellcolor[rgb]{0.867,0.867,0.871}}\textbf{48.07} & 74.68 &  & {\cellcolor[rgb]{0.867,0.867,0.871}}\textbf{55.96} & 80.55 & {\cellcolor[rgb]{0.867,0.867,0.871}}\textbf{50.64} & 78.35 \\
 & \textbf{Product} w/o weight.& {\cellcolor[rgb]{0.867,0.867,0.871}}49.32 & \textbf{74.98} & {\cellcolor[rgb]{0.867,0.867,0.871}}44.4 & \textbf{72.66} &  & {\cellcolor[rgb]{0.867,0.867,0.871}}51.76 & \textbf{83.41} & {\cellcolor[rgb]{0.867,0.867,0.871}}46.97 & 80.92 &  & {\cellcolor[rgb]{0.867,0.867,0.871}}53.99 & 85.47 & {\cellcolor[rgb]{0.867,0.867,0.871}}49.36 & 82.75 \\
 &\textbf{Product} w. weight. & {\cellcolor[rgb]{0.867,0.867,0.871}}\textbf{49.57} & 74.84 & {\cellcolor[rgb]{0.867,0.867,0.871}}\textbf{44.77} & 72.48 &  & {\cellcolor[rgb]{0.867,0.867,0.871}}51.76 & \textbf{83.41} & {\cellcolor[rgb]{0.867,0.867,0.871}}46.97 & \textbf{80.92} &  & {\cellcolor[rgb]{0.867,0.867,0.871}}53.85 & 85.47 & {\cellcolor[rgb]{0.867,0.867,0.871}}49.17 & 82.75 \\ 
\midrule
\multirow{4}{*}{\rotatebox[origin=c]{90}{\textbf{Rank-level}}} & \textbf{Majority} & {\cellcolor[rgb]{0.867,0.867,0.871}}47.81 & 70.49 & {\cellcolor[rgb]{0.867,0.867,0.871}}42.57 & 67.71 &  & {\cellcolor[rgb]{0.867,0.867,0.871}}51.98 & 77.62 & {\cellcolor[rgb]{0.867,0.867,0.871}}46.42 & 75.23 &  & {\cellcolor[rgb]{0.867,0.867,0.871}}54.75 & 80.66 & {\cellcolor[rgb]{0.867,0.867,0.871}}49.91 & 78.53 \\
 & \textbf{Borda count} w/o. weight.& {\cellcolor[rgb]{0.867,0.867,0.871}}44.41 & 73.76 & {\cellcolor[rgb]{0.867,0.867,0.871}}38.72 & 72.11 &  & {\cellcolor[rgb]{0.867,0.867,0.871}}50.65 & 80.5 & {\cellcolor[rgb]{0.867,0.867,0.871}}46.06 & 78.35 &  & {\cellcolor[rgb]{0.867,0.867,0.871}}54.25 & \textbf{85.91} & {\cellcolor[rgb]{0.867,0.867,0.871}}50.09 & \textbf{83.49} \\
 & \textbf{Borda count} w. weight. & {\cellcolor[rgb]{0.867,0.867,0.871}}47.81 & 70.62 & {\cellcolor[rgb]{0.867,0.867,0.871}}42.57 & 67.34 &  & {\cellcolor[rgb]{0.867,0.867,0.871}}51.51 & 77.6 & {\cellcolor[rgb]{0.867,0.867,0.871}}46.06 & 75.05 &  & {\cellcolor[rgb]{0.867,0.867,0.871}}54.53 & 81.06 & {\cellcolor[rgb]{0.867,0.867,0.871}}49.54 & 79.27 \\
 & \textbf{Reciprocal Rank}& {\cellcolor[rgb]{0.867,0.867,0.871}}42.65 & 69.96 & {\cellcolor[rgb]{0.867,0.867,0.871}}37.06 & 66.79 &  & {\cellcolor[rgb]{0.867,0.867,0.871}}48.45 & 81.45 & {\cellcolor[rgb]{0.867,0.867,0.871}}43.3 & 79.27 &  & {\cellcolor[rgb]{0.867,0.867,0.871}}52.58 & 83.76 & {\cellcolor[rgb]{0.867,0.867,0.871}}48.26 & 80.73 \\
\bottomrule
\end{tabular}}
    \caption{Performance of late-fusion methods on \textbf{rare} classes of the Drive\&Act test set}
    \label{tab:rare}
\end{table*}

\mypar{Product-fusion:} The product-fusion $f_{PROD}(\cdot)$ for $N$ classifiers is defined as:
\begin{equation}
    f_{PROD}(c(X))=\gamma\prod^N_{i=1}c(x_i)
\end{equation}
where $\gamma \in \mathbb{R}$ is used as a regularization of the output \cite{masakuna2020performance}.

\mypar{Weighted sum- and product-fusion:} 
Inspired by recent progress of weighted pooling functions~\cite{wang2019comparison}, we further implement variants of sum- and product-fusion, where the individual predictions are weighted via  \textit{Softmax}-normalization amplifying the contribution of the most certain class predictions.
The weighted sum-fusion $f_{WSUM}(\cdot)$ and weighted product-fusion $f_{WPROD}(\cdot)$ for $N$ classifiers are defined as:
{
\small
\begin{equation}
    f_{WSUM}(c(X))=\frac{1}{N} \sum_{i=1}^{N} {w_i c(x_i)}
    \quad 
     f_{WPROD}(c(X))=\gamma\prod^N_{i=1} w_i c(x_i),
\end{equation}}
 \normalsize
where $w_i = \frac{e^{c(x_i)}}{\sum_{j=1}^{N} {e^{c(x_j)}}} $, $i \in {1...N}$.

\subsection{Rank-level fusion}
In contrast to score-level fusion, rank-level fusion leverages the class rankings of multiple classifiers. The magnitude of each class score plays a role only in the ordering of the classes into a ranking list for each classifier. We investigate Majority Voting, the original and weighted Borda Count, as well as Reciprocal Rank Fusion as strategies in this category.

\mypar{Majority Voting:} 
Majority voting first estimates the top-1 predicted behaviour for each individual modality, after which the category, which was predicted by the most unimodal classifiers is selected as the final decision. 
Let $pred_i := \text{argmax}(c(x_i)) \in \mathbb{N}$ be the predicted class from the $i^{\text{th}}$ classifier. The number of the top-1 predictions from all classifiers for class $j$ would then be:
\begin{equation}
    \#j = \#\{pred_i == j | i \in \{1...N\}\}
\end{equation}
where $\#\{\cdot\}$ denotes the set cardinality.
The majority voting $mv(\cdot)$ for $N$ classifiers is defined as:
\begin{equation}
    mv(c(X))=[\{\#1...\#d\}]
\end{equation}

\mypar{Borda Count:}
Another way for combining predictions via late fusion is utilizing a voting system, such as Borda Count \cite{emerson2013original}. The Borda Count voting system is described algorithmically in Algorithm \ref{alg:borda_count}. The class probabilities $c(x_i)$ from all the unimodal models are given as an input. The first loop goes over each of the $N$ classifiers. Their predictions $c(x_i)$ are sorted in descending order so that a ranking list $I$ is created with their indices. In the second loop, the best class prediction for each classifier is given $k$ points, the second-best $k-1$ points, etc., where $k$ is a hyperparameter. This is done for all classifiers, and in the end, these points are added up for the final scoring $\hat{y}$.  

The Borda Count voting resembles a preferential voting system, in contrast to a majoritarian one. This incorporates the uncertainty of each of the separate models’ predictions. In other words, if a model is uncertain about the correct class and ranks it as a second alternative, its prediction would contribute with $k-1$ points for the correct class, instead of 0 points in the case of using a majority vote. However, this relies on the assumption that the classifiers are able to rank the ground truth in their top $k$ predictions, i.e. are not weak.

\begin{algorithm}[!h]
 \KwData{Probability Estimates: $c(x_i) \in \mathbb{R}^{d}$, $i \in \{1...N\}$, where $N=\#$classifiers}
 \KwResult{Fused Class Scores: $\hat{y} \in \mathbb{N}^{d}$}
 $\hat{y} \gets [0...0]$\;
 \For{$i \in \{1...N\}$}{
  $I \gets \textbf{descending\_argsort}(c(x_i))$\;
  \For{$j \in \{k...1\}$}{
     $\hat{y}[I[k-j]] \mathrel{{+}{=}} j$;
  }
 }
 \textbf{return} {$\hat{y}$}\;
 \caption{Borda Count Voting Strategy}
 \label{alg:borda_count}
\end{algorithm}

\mypar{Reciprocal Rank Fusion (RRF):} 
 The $RRF$\cite{cormack2009reciprocal} for $N$ classifiers is defined as:
\begin{equation}
    RRF(c(X)) = [rrf(1)...rrf(d)] 
\end{equation}
where
\begin{equation}
    rrf(j) = \sum_{i=1}^N \frac{1}{m + r_{ij}}
\end{equation}
Cormack et al. \cite{cormack2009reciprocal} introduce the hyperparameter $m\in \mathbb{N}$ and claim that it mitigates the impact of high rankings by outlier systems.

\mypar{Weighted Borda Count:} 
The WBC is an extension of the original algorithm, where the score of each voter is weighted by the corresponding weighting vector $w \in \mathbb{R}^d$. The WBC for $N$ classifiers is defined as:
\begin{equation}
    WBC(c(X))= w \odot BC(c(X)) = w \odot \hat{y} 
\end{equation}
where $\odot$ is the element-wise multiplication operator. The vector $w$ can be computed by an arbitrary weighting function. In our experiments we use the mean softmax outputs, i.e. $w=f_{SUM}(c(X))$. 
We also considered computing the weights via Softmax-normalization over the modalities (as done in the weighted sum- and product-fusion) but observed a significant performance decline.
The reasoning behind $w$ is to enhance the contribution of the most certain class predictions in the fusion stage \cite{drotar2017heterogeneous}.

\section{Experimental Results}

\begin{table*}[]
    \centering
 \scalebox{0.95}{\begin{tabular}{cccccccccccccccc} 
\toprule
 & \multirow{3}{*}{\begin{tabular}[c]{@{}c@{}}\textbf{Fusion}\\\textbf{Method }\end{tabular}} & \multicolumn{4}{c}{\textbf{\#Mod=2 }} &  & \multicolumn{4}{c}{\textbf{\#Mod=4 }} &  & \multicolumn{4}{c}{\textbf{\#Mod=8 }} \\ 
\cline{3-6}\cline{8-11}\cline{13-16}
 &  & \multicolumn{2}{c}{\textbf{Balanced Acc. }} & \multicolumn{2}{c}{\textbf{Unbalanced Acc. }} &  & \multicolumn{2}{c}{\textbf{Balanced Acc. }} & \multicolumn{2}{c}{\textbf{Unbalanced Acc. }} &  & \multicolumn{2}{c}{\textbf{Balanced Acc. }} & \multicolumn{2}{c}{\textbf{Unbalanced Acc.}} \\
 &  & \textbf{\textcolor{red}{Top-1 }} & \textbf{Top-5 } & \textbf{\textcolor{red}{Top-1 }} & \textbf{Top-5} &  & \textbf{\textcolor{red}{Top-1 }} & \textbf{Top-5 } & \textbf{\textcolor{red}{Top-1 }} & \textbf{Top-5} &  & \textbf{\textcolor{red}{Top-1 }} & \textbf{Top-5 } & \textbf{\textcolor{red}{Top-1 }} & \textbf{Top-5 } \\ 
\midrule
\multirow{6}{*}{\rotatebox[origin=c]{90}{\textbf{Score-level}}} &  \textbf{Avg.} w/o weight. (standard) & {\cellcolor[rgb]{0.863,0.863,0.867}}72.68 & 92.23 & {\cellcolor[rgb]{0.863,0.863,0.867}}77.41 & 94.59 &  & {\cellcolor[rgb]{0.863,0.863,0.867}}80.12 & 95.13 & {\cellcolor[rgb]{0.863,0.863,0.867}}84.9 & 96.70 &  & {\cellcolor[rgb]{0.863,0.863,0.867}}82.01 & 96.60 & {\cellcolor[rgb]{0.863,0.863,0.867}}86.27 & 97.54 \\
 & \textbf{Average} w. weight.& {\cellcolor[rgb]{0.863,0.863,0.867}}72.18 & 92.22 & {\cellcolor[rgb]{0.863,0.863,0.867}}76.87 & 94.59 &  & {\cellcolor[rgb]{0.863,0.863,0.867}}80.00 & 95.14 & {\cellcolor[rgb]{0.863,0.863,0.867}}84.67 & 96.70 &  & {\cellcolor[rgb]{0.863,0.863,0.867}}81.96 & 96.64 & {\cellcolor[rgb]{0.863,0.863,0.867}}86.20 & 97.56 \\
 & \textbf{Median} & {\cellcolor[rgb]{0.863,0.863,0.867}}72.68 & 92.23 & {\cellcolor[rgb]{0.863,0.863,0.867}}77.41 & 94.59 &  & {\cellcolor[rgb]{0.863,0.863,0.867}}79.66 & 95.64 & {\cellcolor[rgb]{0.863,0.863,0.867}}84.69 & 97.24 &  & {\cellcolor[rgb]{0.863,0.863,0.867}}81.22 & 96.84 & {\cellcolor[rgb]{0.863,0.863,0.867}}85.98 & 97.71 \\
 & \textbf{Max} & {\cellcolor[rgb]{0.863,0.863,0.867}}71.88 & 92.18 & {\cellcolor[rgb]{0.863,0.863,0.867}}76.55 & 94.51 &  & {\cellcolor[rgb]{0.863,0.863,0.867}}79.28 & 95.06 & {\cellcolor[rgb]{0.863,0.863,0.867}}83.70 & 96.58 &  & {\cellcolor[rgb]{0.863,0.863,0.867}}\textbf{82.76} & 96.47 & {\cellcolor[rgb]{0.863,0.863,0.867}}85.84 & 97.36 \\
 &  \textbf{Product} w/o weight. & {\cellcolor[rgb]{0.863,0.863,0.867}}\textbf{74.51} & \textbf{94.60} & {\cellcolor[rgb]{0.863,0.863,0.867}}\textbf{80.06} & \textbf{96.34} &  & {\cellcolor[rgb]{0.863,0.863,0.867}}80.67 & \textbf{96.54} & {\cellcolor[rgb]{0.863,0.863,0.867}}85.59 & \textbf{97.67} &  & {\cellcolor[rgb]{0.863,0.863,0.867}}82.44 & 97.01 & {\cellcolor[rgb]{0.863,0.863,0.867}}\textbf{86.86} & \textbf{97.92} \\
 &  \textbf{Product} w. weight. & {\cellcolor[rgb]{0.863,0.863,0.867}}74.47 & \textbf{94.60} & {\cellcolor[rgb]{0.863,0.863,0.867}}80.05 & \textbf{96.34} &  & {\cellcolor[rgb]{0.863,0.863,0.867}}\textbf{80.71} & \textbf{96.54} & {\cellcolor[rgb]{0.863,0.863,0.867}}\textbf{85.62} & \textbf{97.67} &  & {\cellcolor[rgb]{0.863,0.863,0.867}}82.44 & 96.99 & {\cellcolor[rgb]{0.863,0.863,0.867}}\textbf{86.86} & 97.90 \\ 
\midrule
\multirow{4}{*}{\rotatebox[origin=c]{90}{\textbf{Rank-level}}} & \textbf{Majority} & {\cellcolor[rgb]{0.863,0.863,0.867}}72.64 & 92.23 & {\cellcolor[rgb]{0.863,0.863,0.867}}77.32 & 94.59 &  & {\cellcolor[rgb]{0.863,0.863,0.867}}79.70 & 95.13 & {\cellcolor[rgb]{0.863,0.863,0.867}}84.62 & 96.70 &  & {\cellcolor[rgb]{0.863,0.863,0.867}}81.51 & 96.51 & {\cellcolor[rgb]{0.863,0.863,0.867}}86.05 & 97.44 \\
 & {\textbf{Borda count}} w/o. weight. & {\cellcolor[rgb]{0.863,0.863,0.867}}65.76 & 92.88 & {\cellcolor[rgb]{0.863,0.863,0.867}}73.95 & 95.03 &  & {\cellcolor[rgb]{0.863,0.863,0.867}}77.83 & 96.18 & {\cellcolor[rgb]{0.863,0.863,0.867}}83.85 & 97.44 &  & {\cellcolor[rgb]{0.863,0.863,0.867}}80.18 & \textbf{97.17} & {\cellcolor[rgb]{0.863,0.863,0.867}}85.64 & 98.08 \\
 & {\textbf{Borda count}} w. weight. & {\cellcolor[rgb]{0.863,0.863,0.867}}72.48 & 92.43 & {\cellcolor[rgb]{0.863,0.863,0.867}}77.23 & 94.66 &  & {\cellcolor[rgb]{0.863,0.863,0.867}}80.11 & 95.14 & {\cellcolor[rgb]{0.863,0.863,0.867}}84.92 & 96.68 &  & {\cellcolor[rgb]{0.863,0.863,0.867}}81.99 & 96.56 & {\cellcolor[rgb]{0.863,0.863,0.867}}86.30 & 97.51 \\
 & \textbf{Reciprocal Rank} & {\cellcolor[rgb]{0.863,0.863,0.867}}65.88 & 92.37 & {\cellcolor[rgb]{0.863,0.863,0.867}}74.08 & 95.29 &  & {\cellcolor[rgb]{0.863,0.863,0.867}}75.36 & 95.37 & {\cellcolor[rgb]{0.863,0.863,0.867}}82.32 & 97.31 &  & {\cellcolor[rgb]{0.863,0.863,0.867}}79.26 & 96.33 & {\cellcolor[rgb]{0.863,0.863,0.867}}85.32 & 97.56 \\
\bottomrule
\end{tabular}}
    \caption{Performance of late-fusion methods on \textbf{common} classes of the Drive\&Act test set}
    \label{tab:common}
\end{table*}

\begin{table*}[]
    \centering
     \scalebox{0.95}{\begin{tabular}{cccccccccccccccc} 
\toprule
 & \multirow{3}{*}{\begin{tabular}[c]{@{}c@{}}\textbf{Fusion}\\\textbf{Method }\end{tabular}} & \multicolumn{4}{c}{\textbf{\#Mod=2 }} &  & \multicolumn{4}{c}{\textbf{\#Mod=4 }} &  & \multicolumn{4}{c}{\textbf{\#Mod=8 }} \\ 
\cline{3-6}\cline{8-11}\cline{13-16}
 &  & \multicolumn{2}{c}{\textbf{Balanced Acc. }} & \multicolumn{2}{c}{\textbf{Unbalanced Acc. }} &  & \multicolumn{2}{c}{\textbf{Balanced Acc. }} & \multicolumn{2}{c}{\textbf{Unbalanced Acc. }} &  & \multicolumn{2}{c}{\textbf{Balanced Acc. }} & \multicolumn{2}{c}{\textbf{Unbalanced Acc.}} \\
 &  & \textbf{\textcolor{red}{Top-1 }} & \textbf{Top-5 } & \textbf{\textcolor{red}{Top-1 }} & \textbf{Top-5} &  & \textbf{\textcolor{red}{Top-1 }} & \textbf{Top-5 } & \textbf{\textcolor{red}{Top-1 }} & \textbf{Top-5} &  & \textbf{\textcolor{red}{Top-1 }} & \textbf{Top-5 } & \textbf{\textcolor{red}{Top-1 }} & \textbf{Top-5 } \\ 
\midrule
\multirow{6}{*}{\rotatebox[origin=c]{90}{\textbf{Score-level}}} & \textbf{Avg.} w/o. weight. (standard)& {\cellcolor[rgb]{0.867,0.867,0.875}}60.25 & 81.36 & {\cellcolor[rgb]{0.867,0.867,0.875}}74.31 & 92.19 &  & {\cellcolor[rgb]{0.867,0.867,0.875}}65.78 & 86.50 & {\cellcolor[rgb]{0.867,0.867,0.875}}81.45 & 94.81 &  & {\cellcolor[rgb]{0.867,0.867,0.875}}68.35 & 88.75 & {\cellcolor[rgb]{0.867,0.867,0.875}}83.01 & 95.87 \\
 & \textbf{Average} w. weight. & {\cellcolor[rgb]{0.867,0.867,0.875}}59.85 & 81.36 & {\cellcolor[rgb]{0.867,0.867,0.875}}73.79 & 92.19 &  & {\cellcolor[rgb]{0.867,0.867,0.875}}65.73 & 86.51 & {\cellcolor[rgb]{0.867,0.867,0.875}}81.25 & 94.81 &  & {\cellcolor[rgb]{0.867,0.867,0.875}}68.46 & 88.7 & {\cellcolor[rgb]{0.867,0.867,0.875}}82.98 & 95.88 \\
 & \textbf{Median} & {\cellcolor[rgb]{0.867,0.867,0.875}}60.25 & 81.36 & {\cellcolor[rgb]{0.867,0.867,0.875}}74.31 & 92.19 &  & {\cellcolor[rgb]{0.867,0.867,0.875}}65.76 & 87.86 & {\cellcolor[rgb]{0.867,0.867,0.875}}81.32 & 95.56 &  & {\cellcolor[rgb]{0.867,0.867,0.875}}67.62 & 90.73 & {\cellcolor[rgb]{0.867,0.867,0.875}}82.74 & 96.29 \\
 & \textbf{Max} & {\cellcolor[rgb]{0.867,0.867,0.875}}59.70 & 81.18 & {\cellcolor[rgb]{0.867,0.867,0.875}}73.49 & 92.06 &  & {\cellcolor[rgb]{0.867,0.867,0.875}}\textbf{66.27} & 86.26 & {\cellcolor[rgb]{0.867,0.867,0.875}}80.53 & 94.63 &  & {\cellcolor[rgb]{0.867,0.867,0.875}}\textbf{69.36} & 88.51 & {\cellcolor[rgb]{0.867,0.867,0.875}}82.70 & 95.67 \\
 & \textbf{Product} w/o weight.& {\cellcolor[rgb]{0.867,0.867,0.875}}61.91 & \textbf{84.79} & {\cellcolor[rgb]{0.867,0.867,0.875}}76.89 & \textbf{94.23} &  & {\cellcolor[rgb]{0.867,0.867,0.875}}66.21 & \textbf{89.97} & {\cellcolor[rgb]{0.867,0.867,0.875}}82.15 & \textbf{96.18} &  & {\cellcolor[rgb]{0.867,0.867,0.875}}68.22 & 91.24 & {\cellcolor[rgb]{0.867,0.867,0.875}}\textbf{83.52} & 96.57 \\
 & \textbf{Product} w. weight.& {\cellcolor[rgb]{0.867,0.867,0.875}}\textbf{62.02} & 84.72 & {\cellcolor[rgb]{0.867,0.867,0.875}}\textbf{76.91} & 94.22 &  & {\cellcolor[rgb]{0.867,0.867,0.875}}66.23 & \textbf{89.97} & {\cellcolor[rgb]{0.867,0.867,0.875}}\textbf{82.18} & \textbf{96.18} &  & {\cellcolor[rgb]{0.867,0.867,0.875}}68.14 & 91.23 & {\cellcolor[rgb]{0.867,0.867,0.875}}83.50 & 96.55 \\ 
\midrule
\multirow{4}{*}{\rotatebox[origin=c]{90}{\textbf{Rank-level}}} & \textbf{Majority} & {\cellcolor[rgb]{0.867,0.867,0.875}}60.23 & 81.36 & {\cellcolor[rgb]{0.867,0.867,0.875}}74.23 & 92.19 &  & {\cellcolor[rgb]{0.867,0.867,0.875}}65.84 & 86.37 & {\cellcolor[rgb]{0.867,0.867,0.875}}81.22 & 94.79 &  & {\cellcolor[rgb]{0.867,0.867,0.875}}68.13 & 88.59 & {\cellcolor[rgb]{0.867,0.867,0.875}}82.84 & 95.75 \\
 & \textbf{Borda count} w/o weight. & {\cellcolor[rgb]{0.867,0.867,0.875}}55.08 & 83.32 & {\cellcolor[rgb]{0.867,0.867,0.875}}70.81 & 92.99 &  & {\cellcolor[rgb]{0.867,0.867,0.875}}64.24 & 88.34 & {\cellcolor[rgb]{0.867,0.867,0.875}}80.48 & 95.74 &  & {\cellcolor[rgb]{0.867,0.867,0.875}}67.22 & \textbf{91.54} & {\cellcolor[rgb]{0.867,0.867,0.875}}82.48 & \textbf{96.78} \\
 & \textbf{Borda count} w. weight.& {\cellcolor[rgb]{0.867,0.867,0.875}}60.15 & 81.53 & {\cellcolor[rgb]{0.867,0.867,0.875}}74.15 & 92.23 &  & {\cellcolor[rgb]{0.867,0.867,0.875}}65.81 & 86.37 & {\cellcolor[rgb]{0.867,0.867,0.875}}81.46 & 94.76 &  & {\cellcolor[rgb]{0.867,0.867,0.875}}68.26 & 88.81 & {\cellcolor[rgb]{0.867,0.867,0.875}}83.03 & 95.88 \\
 & \textbf{Reciprocal Rank}& {\cellcolor[rgb]{0.867,0.867,0.875}}54.26 & 81.17 & {\cellcolor[rgb]{0.867,0.867,0.875}}70.78 & 92.75 &  & {\cellcolor[rgb]{0.867,0.867,0.875}}61.91 & 88.41 & {\cellcolor[rgb]{0.867,0.867,0.875}}78.85 & 95.70 &  & {\cellcolor[rgb]{0.867,0.867,0.875}}65.92 & 90.04 & {\cellcolor[rgb]{0.867,0.867,0.875}}82.02 & 96.06 \\
\bottomrule
\end{tabular}}
    \caption{Performance of late-fusion methods on \textbf{all} classes of the Drive\&Act test set}
    \label{tab:all}
\end{table*}

\subsection{Testbed} 
We chose the multimodal Drive\&Act dataset~\cite{MartinRoitberg2019} as our evaluation testbed as it provides a diverse set of driver behaviours recorded with eight  synchronized sensors, therefore enabling a comprehensive study of fusion techniques with a large set of modalities. 
Drive\&Act modalities include one RGB-, one depth-, and six Near-Infrared (NIR) views with 12 hours recorded in total. 
The videos are labeled with a hierarchical annotation scheme, where 34 fine-grained activities constitute the main evaluation level.
We follow the original evaluation protocol comprising three splits into \textit{training}, \textit{validation} and \textit{test}  with no intersection of  drivers (10, 2 and 3 people respectively).

The 34 fine-grained activity classes of Drive\&Act are unbalanced: the number of examples per behaviour type ranges from $19$ (\textsl{taking laptop from backpack}) to $2797$ (\textsl{sitting still}). Since machine learning models rely strongly on the amount of training data,  we report the performance separately for \textit{common}, \textit{rare}, and \textit{all} categories, as suggested in ~\cite{roitberg2020cnn}.
We report the top-1 and top-5 accuracies under balanced and unbalanced conditions. For the balanced accuracy, the metric is computed individually for each class and  the average  over all $34$ behaviours is reported. 
The unbalanced accuracy is the  percentage of correctly recognized examples over the complete dataset, (\ie, in unbalanced settings the underrepresented classes acquire a smaller weight).
The additional top-5 accuracy is especially useful on Drive\&Act since  we might be interested in coarser recognition and dismiss mistakes caused by  highly
similar classes (such as \textsl{opening} and \textsl{closing bottle}). 

We use $k=5$, $m=60$ and $\gamma = 1$ for Borda Count and Reciprocal Rank Fusion and product fusion according to the previous literature~\cite{cormack2009reciprocal,masakuna2020performance}.
For training the eight unimodal classifiers, the initial I3D weights are initialized using the Kinetics dataset~\cite{carreira2017quo}, as done in the original Drive\&Act work~\cite{MartinRoitberg2019} and then optimized for driver behaviour classification with stochastic gradient descent using the initial learning of 0.01   (decreased by a factor  of 10 after 50 and 100 epochs), momentum of
0.9, weight decay of 1e-7 and  mini-batch size of 8. During training, temporal data augmentation samples clips of
64 frames and spacial data augmentation computes random
crops of size 224 × 224.

\subsection{Results} 
The main objective of our experiments is to determine the impact of fusion strategies for the probability estimates of multimodal predictors in the context of driver observation, where averaging has presumably been the most common choice for fusion at decision-level~\cite{MartinRoitberg2019, kopuklu2021driver, martin2018body, khan2021modified}.
Tables \ref{tab:rare}, \ref{tab:common} and \ref{tab:all} display balanced and unbalanced top-1 and top-5 accuracies for different fusion schemes and \emph{rare}, \emph{common} and \emph{all} driver behaviour categories respectively. 
In all settings, we consider $2$, $4$ and all $8$ Drive\&Act modalities (the $2$ and $4$ modalities were chosen by selecting the first $2/4$ modalities from a random permutation of all available views). 
In Table \ref{tab:rare} (underrepresented behaviours), product-fusion and max-fusion yielded the best outcome (for example, $1.82\%$, $5.53\%$, $2.01\%$ and $5.51\%$ gain in performance compared to the conventional score averaging for the different metrics and four modalities). 
Interestingly, the models with best results in terms of the top-1 accuracy are not necessarily the best as it comes to the top-5 results. 
This hints that some models are better at coarser recognition, since the top-5 metrics often omits fine-grained confusions, such as \textsl{preparing food} vs. \textsl{eating}. 
For instance, Borda Count is the best performing fusion method for $8$ modalities in terms of the top-5 accuracy, while it usually yields similar or slightly worse results compared to averaging looking at the top-1 metrics.
While additional weighting does not have a significant influence on product- and average-fusion, it positively impacts the Borda Count results. 

\begin{figure*}
    \centering
    \includegraphics[width=\linewidth]{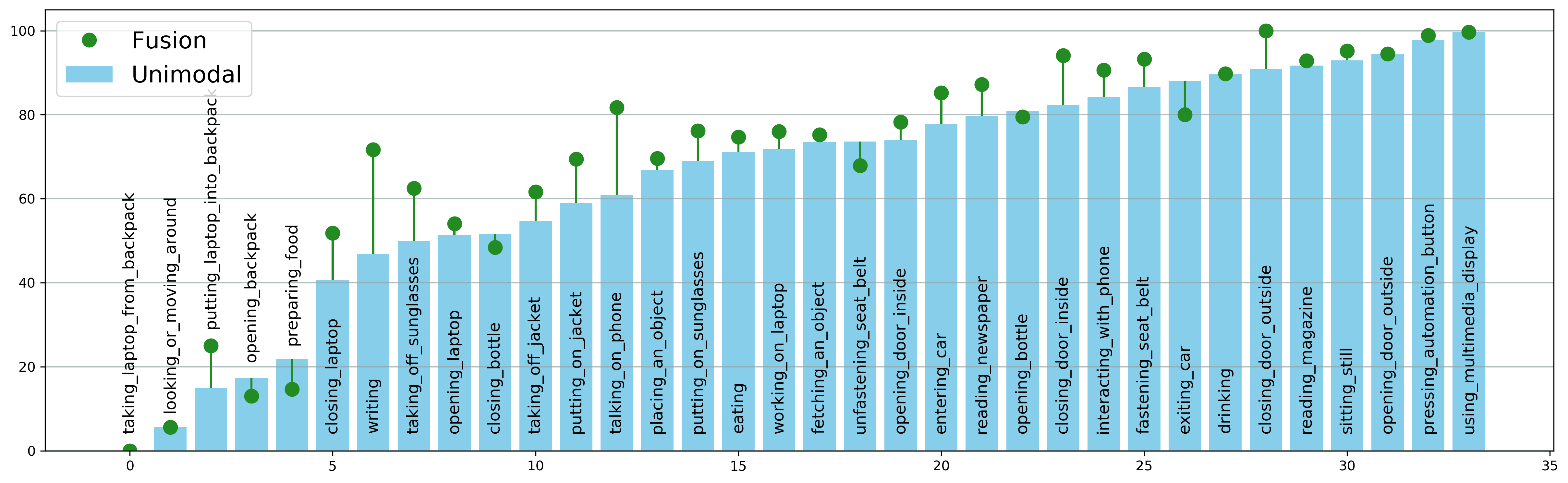}
    \caption{Per-category accuracy for
the best  unimodal classifier (blue bar) and  a multimodal model with eight views and product-fusion (green dot).}
    \label{fig:recall_comprison}
\end{figure*}

\begin{table}
\centering
\scalebox{0.93}{\begin{tabular}{lcccc} 
\toprule
\multirow{2}{*}{\textbf{ Modality }} & \multicolumn{2}{c}{\textbf{Balanced Acc. }} & \multicolumn{2}{c}{\textbf{Unbalanced Acc.}} \\
 & \textcolor{red}{\textbf{Top-1 }} & \textbf{Top-5 } & \textcolor{red}{\textbf{Top-1 }} & \textbf{Top-5} \\ 
\midrule
Center mirror, NIR & {\cellcolor[rgb]{0.871,0.871,0.878}}63.09 & \textbf{88.60} & {\cellcolor[rgb]{0.871,0.871,0.878}}77.80 & \textbf{94.63 } \\
A-Column driver, NIR & {\cellcolor[rgb]{0.871,0.871,0.878}}59.92 & 87.19 & {\cellcolor[rgb]{0.871,0.871,0.878}}73.69 & 94.09 \\
Face view, NIR& {\cellcolor[rgb]{0.871,0.871,0.878}}42.32 & 70.23 & {\cellcolor[rgb]{0.871,0.871,0.878}}55.74 & 84.84 \\
Ceiling (back view), NIR& {\cellcolor[rgb]{0.871,0.871,0.878}}61.87 & 84.18 & {\cellcolor[rgb]{0.871,0.871,0.878}}76.84 & 93.03 \\
A-Column co-driver, NIR & {\cellcolor[rgb]{0.871,0.871,0.878}}\textbf{65.05 } & 87.52 & {\cellcolor[rgb]{0.871,0.871,0.878}}\textbf{78.59 } & 94.38 \\
A-Column co-driver, RGB & {\cellcolor[rgb]{0.871,0.871,0.878}}62.70 & 84.52 & {\cellcolor[rgb]{0.871,0.871,0.878}}74.80 & 92.91 \\
 A-Column co-driver, Depth& {\cellcolor[rgb]{0.871,0.871,0.878}}59.83 & 84.41 & {\cellcolor[rgb]{0.871,0.871,0.878}}71.73 & 92.47 \\
\midrule
Multimodal (product) & {\cellcolor[rgb]{0.871,0.871,0.878}}68.22 & 91.24 & {\cellcolor[rgb]{0.871,0.871,0.878}}83.52 & 96.57\\
\bottomrule
\end{tabular}}
    \caption{Unimodal performance for all classes in Drive\&Act }
    \label{tab:unimodal}
\end{table}

These results are confirmed through our experiments on \emph{common} and \emph{all} categories (Tables \ref{tab:common} and \ref{tab:all}): product- and max-fusion alternate in being the frontrunner, while averaging is not the most effective choice in all settings.
Interestingly, while max-fusion oftentimes outperformed product-fusion by a small margin, product-fusion is consistently more effective as it comes to top-5 accuracy, indicating, that it might be useful in coarser recognition. 
Overall, score-level approaches suit better than ranking-based strategies (with very few exceptions, where Borda Count is effective in terms of the top-5 accuracy).

As expected, utilizing more modalities positively impacts the recognition rates (for example, we achieve the top-1 balanced accuracy of $61.91\%$, $66.21\%$ and $68.22\%$ for $2$, $4$, and $8$ modalities and all categories, see Table \ref{tab:all}). 
As previously mentioned, the modality choice was conducted via a random permutation of all Drive\&Act data sources. 
Since the first  modality in the resulting sequence was \emph{A column co-driver, depth}, adding $1$, $3$ and $7$ additional modalities improves the unimodal performance by $2.1\%$, $6.38\%$ and $8.39\%$ accordingly (see Table \ref{tab:unimodal} for the unimodal results). 
Lastly, in Figure \ref{fig:recall_comprison} we  compare our multimodal system (eight modalities with product-fusion) to the best performing unimodal view, which is \emph{A column co-driver, NIR} according to Table \ref{tab:unimodal}.
The individual categories in Figure \ref{fig:recall_comprison} are sorted by their accuracy in the unimodal setting, giving insight on how hard-to-recognize these behaviour types are.
Overall, multimodality leads to performance improvement in almost all behaviour types, but the effect is different depending on the visibility and recognition difficulty: the largest benefits of multimodality were observed in driver behaviours with medium recognition difficulty.
For instance, classification of examples with the driver \textsl{writing}, \textsl{taking off sunglasses} or \textsl{talking on phone} was improved by $24.86\%$, $12.5\%$ and $20.81\%$. 
For ``easier'' driver behaviours, using more modalities positively influenced the performance but the effect is rather small (for example, only $2.24\%$ improvement for sitting still).
This is not surprising, as one effective modality might be already sufficient to recognize such activities. 
Interestingly, the results were rather mixed for very ``hard to recognize'' driver states, as the performance is improved in some cases ($10\%$ increase for \textsl{putting laptop into backpack} but $4\%$ and $7\%$ decline for \textsl{opening backpack} and \textsl{preparing food}, which is often confused with \textsl{eating}).
Since we considered the best performing unimodal classifier, we believe that for certain difficult categories this modality was overwhelmingly better than other sensors, which rather constituted additional noise. 
The choice of modalities should therefore depend on the recognition use-case and behaviours-of-interest, but if a broad range of diverse secondary driver behaviours is required, multimodality is a powerful tool as it complements the advantages and unique characteristics  of the individual sensors.

\section{Conclusion}

In this work, we revisit the paradigm of decision-level fusion in the context of multimodal driver observation, where the predictions of the individual unimodal classifiers were predominantly joined via score averaging in the past~\cite{MartinRoitberg2019, kopuklu2021driver, martin2018body, khan2021modified}. 
We operationalize and study different variants of seven decision-level fusion paradigms used in general machine learning literature in the context of driver behaviour understanding.
We train eight unimodal classifiers on data provided by eight different cameras placed inside the vehicle cabin using a standard backbone neural network for driver activity categorization and equip them with different types of decision-level fusion modules for linking the probability estimates in a final decision.
We found that late fusion based on the product-rule and max-rule lead to the best recognition results, but the effect depends on the task difficulty and number of modalities. 
This suggests that while the selection of the fusion scheme impacts the driver activity recognition performance noticeably, the conventional strategy of averaging the prediction scores is usually not the best choice.

\mypar{Acknowledgements}
This work was partially supported by the Competence Center Karlsruhe for AI Systems Engineering (CC-KING) sponsored by the Ministry of Economic Affairs, Labour and Housing Baden-Württemberg.

{\small
	\bibliographystyle{IEEEtran}
	\bibliography{egbib,score_fusion_learned,score_fusion_rule,score_fusion_driver}
}

\end{document}